\documentclass{article}
\usepackage{spconf,amsmath,graphicx}
\usepackage{amsfonts}
\usepackage{color}
\usepackage[algoruled,boxed,lined]{algorithm2e}
\usepackage{amsmath}
\usepackage{float}
\usepackage[table]{xcolor}

\DeclareMathOperator*{\argmin}{arg\,min}

\title{Multiple Subspace Alignment improves Domain Adaptation}
\name{Kowshik Thopalli\textsuperscript{1}, Rushil Anirudh\textsuperscript{2}, Jayaraman J. Thiagarajan\textsuperscript{2}, Pavan Turaga\textsuperscript{1} \thanks{This work was performed under the auspices of the U.S. Department of Energy by Lawrence Livermore National Laboratory under Contract DE-AC52-07NA27344. KT and PT were supported in part by ARO grant number W911NF-17-1-0293 and NSF CAREER award 1451263.
}}
\address{\textsuperscript{1} Geometric Media Lab, \enspace     \textsuperscript{2} Center for Applied Scientific Computing,\\Arizona State University,\enspace Lawrence Livermore National Laboratory 
\\Email: \{kthopall@asu.edu,\:anirudh1@llnl.gov,\: jjayaram@llnl.gov, \: pturaga@asu.edu\} }

\begin{document}
%
\maketitle
\begin{abstract}
 We present a novel unsupervised domain adaptation (DA) method for cross-domain visual recognition. Though subspace methods have found success in DA, their performance is often limited due to the assumption of approximating an entire dataset using a single low-dimensional subspace. Instead, we develop a method to effectively represent the source and target datasets via a collection of low-dimensional subspaces, and subsequently align them by exploiting the natural geometry of the space of subspaces, on the Grassmann manifold. We demonstrate the effectiveness of this approach, using empirical studies on two widely used benchmarks, with state of the art domain adaptation performance.

\end{abstract}
\section{Introduction}

\label{sec:intro}
Powered by advances in representation learning, the field of computer vision has witnessed unprecedented growth in predictive modeling. Despite the effectiveness of these solutions, it is well known that their performance suffers when the trained models are deployed in new environments, particularly when there are systematic domain shifts not known \emph{a priori}~\cite{Ben-David:2010:TLD:1745449.1745461}. For example, a classifier trained exclusively using images from a DSLR camera might not produce precise predictions on target images from a mobile phone camera. Tackling this discrepancy is central to enabling the effective use of such predictive models in practice. Consequently, a large class of techniques that can leverage labeled data in a source domain to infer an accurate model in an unobserved and unlabeled target domain, have been developed~\cite{Gong2012GeodesicFK},\cite{Gopalan2011DomainAF}. Commonly referred to as \textit{unsupervised domain adaptation}, these techniques attempt to learn domain-invariant representations and subsequently, align the two datasets by minimizing some form of divergence in the learned feature space.
\vspace{5pt}

\noindent Broadly, domain adaptation methods can be categorized as \textit{conservative}, which assume that there exists a single classifier in the hypothesis space that performs well in both domains, or \textit{non-conservative}, which avoid that assumption \cite{shu2018a}. Regardless of the category, the methods vary based on their choice of representation and the strategy to align the source and target distributions. For example, optimal transport has proven to be a powerful alignment strategy regardless of how the datasets are represented \cite{seguy2018large,Courty2017OptimalTF}. State-of-the-art techniques adopt domain adversarial training \cite{tzeng2017adversarial,DBLP:conf/icml/HoffmanTPZISED18}, which
attempts to construct a feature learner that matches the source and target distributions in a latent space. However, these methods assume access to both source and target datasets at train time, which can be unrealistic in several scenarios. 
\vspace{5pt}

\noindent In this paper, we consider the problem where we do not have access to labeled target data, while learning the representations. Instead of inferring features based on an alignment objective (e.g. adversarial loss), the two steps of representation learning and alignment are decoupled. More specifically, our approach performs subspace modeling to build data representations, and  utilizes tools from differential geometry of the Grassmann manifold to perform alignment. This general idea has been explored by several works in the past \cite{Fernando2014SubspaceAF,DBLP:conf/iccv/FernandoHST13, Gong2012GeodesicFK,sun2015subspace}; however, those methods rely on a simplistic model, a single low-dimensional subspace, to model datasets. 
\vspace{5pt}

\begin{figure}[t]
  \centering
  \centerline{\includegraphics[clip,trim={0 120 5 120},width=\columnwidth]{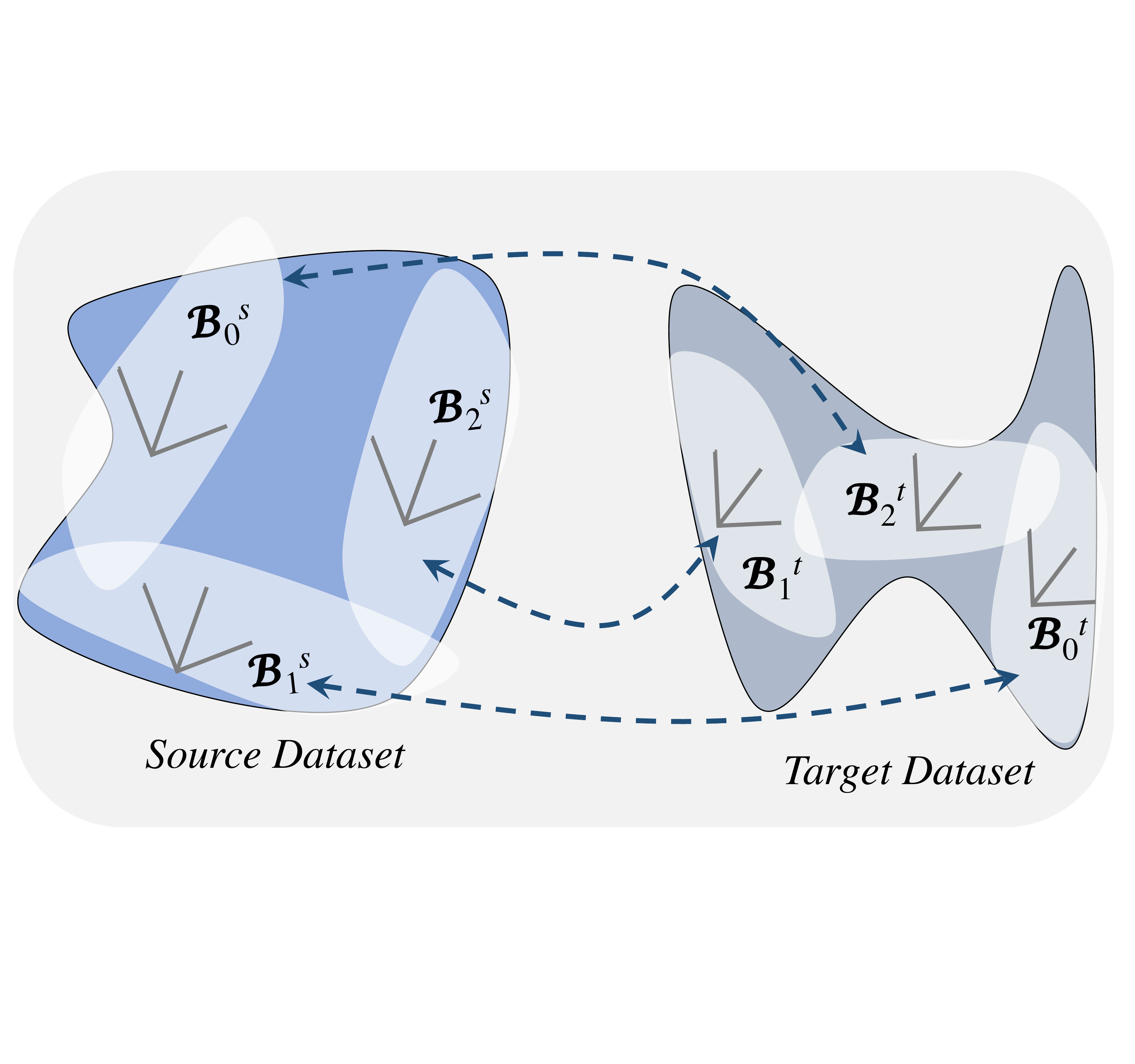}}
  \vspace{-1cm}
  
\caption{\small{\textbf{Multi-subspace alignment:} Our approach proposed to represent the source and target datasets via a collection of discriminative subspaces. We also solve for the correspondance between the source and target subspaces by exploiting the natural geometry of subspaces on the Grassmann manifold. The correspondance is shown here by dashed lines.}}
\label{fig:proposed}
\vspace{-15pt}
\end{figure}

\noindent In contrast, we show that complex datasets can be more effectively modeled through a union of multiple low dimensional subspaces, and this can thereby lead to improved alignment even under challenging domain shifts. A related idea has been explored in \cite{shrivastava2014unsupervised}, wherein they fit multiple subspaces to clusters in the two datasets, and align via parallel transport with regard to a single mean subspace. When compared to our approach, this does not guarantee that the subspaces are discriminative and the alignment fails to take advantage of the increased representational capacity. We on the other hand develop a simple, yet effective, strategy for fitting multiple low-dimensional subspaces to the two domains, and propose to align them by exploiting the natural geometry of linear subspaces on the Grassmann manifold, in order to construct domain-invariant features. Our approach falls under the \textit{conservative} category of domain adaptation. When compared to domain adversarial training, a crucial advantage of our approach is that it is applicable even with smaller datasets. An overview of the proposed approach is shown in fig \ref{fig:proposed}.\vspace{5pt}

\noindent We perform experiments on all benchmarks from the  Office-Caltech10 dataset, using both pre-defined SURF descriptors as well as features extracted from a pre-trained deep neural network architecture. Our results show that the proposed approach achieves state-of-the-art performance on this benchmark, outperforming several existing methods for unsupervised adaptation. More importantly, we observe that the multiple subspace assumption provides significant performance improvements over approaches that use a single subspace.

 \vspace{-10pt}

\section{Preliminaries}
\label{sec:basics}
In this section, we briefly review the mathematical foundations used in designing our approach. 
\vspace{10pt}

\noindent \textbf{Grassmann manifold:} The core idea of our approach relies on representing a dataset with a collection of linear subspaces -- which are known to be points on the Grassmann manifold,  $\mathbb{G}_{r,n}$, which describes the collection of all $r$-dimensional linear subspaces in $\mathbb{R}^n$. Each point on $\mathbb{G}_{r,n}$ is represented as a basis set, \textit{i.e.} a collection of $r$ orthonormal vectors $\mathcal{B} := \{b_1, b_2,\dots, b_r\}$. For more details on the geometry and properties of the manifold, we refer our readers to more comprehensive sources \cite{edelman1998geometry,absil2009optimization}. 
\\

\noindent \textbf{Metric on the Grassmann manifold:} We use the symmetric directional distance $(d_{\Delta})$ to compute distances between points on the Grassmann manifold. We show a more general form of the distance function, which applies to even subspaces with different dimensions $r$ \cite{sun2007further}. This is a popularly adopted measure in several applications~\cite{basri2011approximate, sharafuddin2010know}. Formally, this can be stated as follows:
\begin{equation}
\begin{aligned}
\label{eq:chord}
d_{\Delta}(\mathcal{B}_1,\mathcal{B}_2) =  \left(\textrm{max}(r{_1},r_2)-\sum_{i,j=1}^{r_1,r_2}({b_{1,i}}^{\textrm{T}}b_{2,j})^2\right)^{\frac{1}{2}},
\end{aligned}   
\end{equation}where, $r_1$ and $r_2$ denote the ranks of the basis sets $\mathcal{B}_1$ and $\mathcal{B}_2$ respectively. Note, this expression is equivalent to the chordal metric \cite{ye2016schubert} and is applicable to the case when $r_1 = r_2$.

\vspace{10pt}


\section{Approach: Multi-Subspace Alignment}
\label{sec:Methods}
In this section, we describe the proposed approach for domain adaptation, which first builds a multi-subspace representations for the two domains, and then performs alignment through Grassmann analysis. We assume we have access to a labeled source dataset $\{X^s_i, y^s_i\}_{i=1}^{N_s} \subset D_S$, and an unlabeled target dataset $\{X^t_i\}_{i=1}^{N_t} \subset D_T$, where $D_S$ and $D_T$ denote the source and target domains, $N_s,N_t$ are the number of samples. We assume that every sample in both the domains, $X \in \mathbb{R}^d$, but drawn from different, unknown distributions. Our goal in this paper is to transform the source dataset in such a way that a classifier trained only on the adapted source dataset generalizes well to the unlabeled target dataset. 

\subsection{Fitting multiple subspaces}
The first step in our approach is to learn effective representations for the source and target datasets independently, such that aligning those features will lead to maximal knowledge transfer across the domains. To this end, we develop a greedy approach that identifies a set of low-dimensional subspaces, that can collectively describe all variations in the high-dimensional dataset. Following the standard practice in subspace methods for domain adaptation, we begin by fitting a $k-$dimensional subspace using Principal Component Analysis (PCA) on the entire dataset. Subsequently, we select samples that incur high errors, measured through a normalized reconstruction error for each sample and compared against a user-defined tolerance $\tau$, in the subspace approximation. In order to improve the approximation, we fit a second subspace that produces a high-fidelity representation for the selected samples. We continue this process until the following criteria is met: total number of samples with errors above $\tau$ is less than the chosen subspace dimension $k$




\begin{algorithm}[t]
	\caption{Dataset Approximation using Multiple Subspaces}
	\DontPrintSemicolon
	\SetAlgoLined
	{{\textbf{Input}: \small{Dataset -- $\mathbf{X}$, subspace dimension -- $k$, tolerance $\tau$} }}
	\BlankLine
	{{\textbf{Output}: Collection of subspaces ($\mathcal{M}$), list of subspace IDs to which each sample is associated ($P$)}}
	\BlankLine
	Initialize $\mathcal{M}$ as an empty set\;

	sID $= 1$\;
	\While{no convergence}{
		$ \mathcal{B} \leftarrow$ PCA($\mathbf{X}$, k) // fit subspace\; 
		$ids \leftarrow \{i \in [N]|X_i \in \mathbf{X}, \mbox{error}(X_i, \mathcal{B})\geq\tau\}$\;
		$\mathbf{X}_e \leftarrow \mathbf{X}[ids]$ // errors \;
		$\mathbf{X}_v \leftarrow \mathbf{X}\setminus \mathbf{X}_e$ // inliers \; 
		$\mathcal{B} \leftarrow$ PCA($\mathbf{X}_v$, k) // refit subspace\;		
		$P[i] =$ sID, $\forall i \in ids$\;					
		$\mathcal{M}\leftarrow \mathcal{M} \cup \mathcal{B}$\;
		$\mathbf{X} \leftarrow \mathbf{X}_e$\;	
		sID = sID $+ 1$
	}

	\label{alg:alg1}
\end{algorithm}
%
%
%
%
%
%
%
%
%
%
%
\vspace{-2pt}
\subsection{Alignment of multiple subspaces}
Using Algorithm \ref{alg:alg1}, the source and target datasets are now represented via collection of subspaces and are denoted by $\mathcal{M}_S$ and $\mathcal{M}_T$ respectively. The next crucial step in achieving domain adaptation is aligning the set of source subspaces with those from the target dataset. Domain alignment is a well-studied problem, and hence there exist a wide range of strategies to solve the alignment problem;, e.g. dynamic programming and  optimal transport; however, most of these methods operate directly on the pairwise distance matrix between samples across the domains. However, in our current setup, since the samples that need to be aligned are points on the Grassmann manifold, we propose to utilize the distance function defined in \eqref{eq:chord}. Note, when there is a single low-dimensional subspace in each domain, approaches such as the geodesic flow kernel can be used~\cite{Gong2012GeodesicFK}. However, no such techniques exist for dealing with multi-subspace representations. Hence, we resort to a greedy strategy that assigns unique matches to each source subspace by finding its closest match in the set of target subspaces. This is carried out using geometric distances computed between every pair of subspaces in $\mathcal{M}_S$ and $\mathcal{M}_T$. Naturally, the complexity of this step grows quadratically with the number of subspaces and scalability can be a bottleneck with large datasets. In such cases, a more efficient alignment strategy such as stochastic optimal transport \cite{seguy2018large} could be used as an alternative.


\vspace{10pt}

\noindent \textbf{Alignment between Matched Subspaces: }We use the subspace alignment (SA) strategy proposed in \cite{DBLP:conf/iccv/FernandoHST13} to align each source subspace with its the matched target subspace. The alignment of two subspaces on the Grassmann manifold is modeled via a linear transformation, which can be computed in closed form \cite{DBLP:conf/iccv/FernandoHST13}. In other words, the basis vectors are aligned using a transformation matrix $\mathbf{A} \in \mathbb{R} ^{k \times k}$ from the $m^{\text{th}}$ source subspace $\mathcal{B}_m^s$ to the $n^{\text{th}}$ target subspace $\mathcal{B}_n^t$. The transformation is obtained by minimizing the following objective: 
\begin{align}
\label{eq:objective}
\begin{split}
\mathbf{A}^* = \argmin_\mathbf{A} \left\|\mathcal{B}_m^s \mathbf{A} - \mathcal{B}_n^t\right\|_F,
\end{split}
\end{align}where $\|.\|_F$ denotes the Frobenius norm. The solution to \eqref{eq:objective} is given as $\mathbf{A}^* = (\mathcal{B}_m^s)^T \mathcal{B}_n^t$. This implies that the adjusted coordinate system, also referred as the \textit{target aligned source coordinate  system} can be obtained as \begin{equation}
\mathcal{B}_m^{ta} =  \mathcal{B}_m^s (\mathcal{B}_m^s)^T \mathcal{B}_n^t.
\label{eqn:align}
\end{equation}We obtain representations for the source dataset by projecting each sample $\{X_i^s\}$ onto the corresponding target aligned subspace as indicated by the association list $P_S[i]$. On the other hand, each of the target samples $\{X_i^t\}$ are directly projected onto the target subspace $\mathcal{B}_n^t$, where $P_T[i] = n$. Following previous works such as \cite{sun2015subspace,DBLP:conf/iccv/FernandoHST13}, a nearest neighbor (NN) classifier is then used with these representations, which we expect to be effective for both source and target datasets. An overview of this algorithm can be found in Algorithm \ref{alg:proposed}.

\begin{algorithm}[t]
\caption{Proposed domain adaptation algorithm}
\SetAlgoLined
{{\textbf{Input}: \small{Source dataset $\mathbf{X}^s$, target dataset $\mathbf{X}^t$, source labels $\mathbf{y}^s$, subspace 
			dimension $k$, tolerance parameters $\tau^s$ and $\tau^t$}}}
\BlankLine
{{\textbf{Output}: Predicted target labels $\{y_j^t\}$}}
\BlankLine

$\mathcal{M}_S, P_S \leftarrow$ fit($\mathbf{X}^s, k, \tau^s$) // Invoke Algorithm 1 \;

$\mathcal{M}_T, P_T \leftarrow$ fit($\mathbf{X}^t, k, \tau^t$) // Invoke Algorithm 1 \;

$D \leftarrow$ distancematrix($\mathcal{M}_S,\mathcal{M}_T$) // using \eqref{eq:chord} \;

Greedy matching of subspaces across domains with $D$ \;

Compute $\mathcal{B}_m^{ta}$, for every $\mathcal{B}_m^{s} \in \mathcal{M}_S$ // using \eqref{eqn:align} \;

Obtain features $Z_i^s = \mathcal{B}_m^{ta} X_i^s$, where $P_S[i] = m$ \;

Obtain features $Z_j^t = \mathcal{B}_n^t X_j^t$, where $P_T[j] = n, \forall i$ \;

$\{y_j^t\} \leftarrow$ NN-Classifier$(\{Z_i^s, y_i^s\}, \{Z_j^t\}), \forall j$ \;
\label{alg:proposed}

\end{algorithm}

\vspace{-5pt}
\section{Experiments}
\label{sec:experiments}
\vspace{-2pt}
We evaluate our approach on benchmark domain adaptation tasks from widely used object recognition datasets, similar to several existing works\cite{Gong2012GeodesicFK}, \cite{Fernando2014SubspaceAF},  \cite{sun2015subspace} and \cite{ShrivastavaWACV2014}. 
\vspace{-0.4cm}
\subsection{Datasets}
\begin{figure}[H]
  \centering
  \centerline{\includegraphics[width=8.5cm]{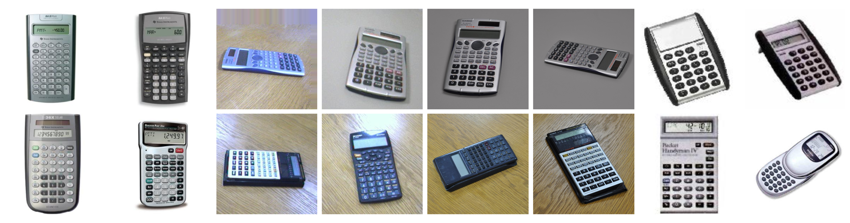}}
  \vspace{-10pt}
  \caption{Example images from Office-Caltech10. From left to right- Amazon, Webcam, DSLR and Caltech domains}
\label{fig:office_caltech}
\vspace{-10pt}
\end{figure}

The Office-Caltech10 \cite{Gopalan2011DomainAF,Saenko2010AdaptingVC} dataset is comprised of images from four different domains - Amazon (online merchant), Webcam (acquired using a webcam), DSLR (captured through a digital SLR camera) and Caltech (from Caltech dataset). For brevity, we refer to these domains using the letters A, W, D, and C respectively. The variations across the domains can be attributed to a variety of factors including presence/absence of background, lighting conditions, noise, etc. Each of the domains is comprised of images from $10$ common object categories (backpack, bike, calculator, headphones, keyboard, laptop computer, monitor, mouse, mug, and projector). Sample images for the \textit{Calculator} class from each of the four domains are shown in Figure \ref{fig:office_caltech}. We consider adapting between every pair of domains, resulting in a total of $12$ cases.

We consider the use of two feature sets, commonly adopted in vision systems, for each of these domains: (i) SURF \cite{bay2006surf} features: Each image is encoded using a visual dictionary of 800 words, and the encodings are normalized to have a zero mean and unit standard deviation in each dimension, (ii) Decaf representations \cite{Donahue2013DeCAFAD}: Following \cite{Donahue2013DeCAFAD}, we construct features using the activations ($4096$ dimensions) from the sixth layer of a network pre-trained on the ImageNet dataset \cite{deng2009imagenet}. 

\begin{table}[t]
\begin{center}
\begingroup
\setlength{\tabcolsep}{10pt} 
\renewcommand{\arraystretch}{1.2}
\resizebox{\columnwidth}{!}{%
\begin{tabular}{|c|c|c|c|c|c|c|}
\hline
\cellcolor{gray!25}{\textbf{Domains}} & \cellcolor{gray!25}\textbf{NA} & \cellcolor{gray!25}\textbf{GFK}& \cellcolor{gray!25}\textbf{SA}& \cellcolor{gray!25}\textbf{SDA} & \cellcolor{gray!25}\textbf{PT} & \cellcolor{gray!25}\textbf{Proposed} \\
\hline
A-C & 26.03 & 35.6 & 38.4 & 39.98 & 41.4 &\textbf{ 44.52} \\

C-A & 23.69 & 36.9 & 40.6 & 49.69 & 49.4 & \textbf{57.09}\\

D- A & 28.57 & 32.5 & 33.12 & 38.73 & 38.2 & \textbf{45.92} \\

A-D & 25.48 & 35.2 & 37.57 & 33.76 & -- &\textbf{ 41.4} \\

W-A & 20.8 & 31.1 & 37.4 & 39.25 & 38.7 & \textbf{49.37} \\

A-W & 23.3 & 34.4 & 38.6 & 30.85 & 40.4 & \textbf{43.05} \\

W-C & 16.4 & 27.2 & 32.3 & 34.73 & \textbf{37.1} & 36.51 \\

C-W & 20 & 33.7 & 36.8 & 38.98 & & \textbf{40.67 }\\

W-D & 40.5 & 70.6 & 80.3 & 75.8 & & \textbf{85.98} \\

D-W & 53 & 74.9 & 83.6 & 76.95 & 84.8 & \textbf{88.81} \\

C-D & 21.7 & 35.2 & 39.6 & 40.13 & \textbf{48.2} & 47.13 \\

D-C & 24.8 & 29.8 & 32.4 & \textbf{35.89} & -- & 35.4 \\
\hline
\hline
Avg. & 27.07 & 39.8 & 44.24 & 44.56 & 47.12 &\textbf{ 51.24 }\\
\hline

\end{tabular} %
}
\vspace{-20pt}
\endgroup

\end{center}
\caption{Classification accuracy on Office-Caltech dataset with SURF features}
\vspace{-10pt}
\label{table:tab_Surf}
\end{table}

\subsection{Experimental Setup}
We compare our method with the following baseline methods in the unsupervised domain adaptation literature. As described earlier, we use a nearest neighbor classifier in all of our experiments.

\noindent \textbf{NA} No Adaptation - In this na\"ive baseline, we compare the performance of classifiers from the source domain applied directly to the target domain.

\noindent \textbf{GFK} Geodesic flow kernel\cite{Gong2012GeodesicFK} - This approach fits a single subspace to each domain and considers all the points along the geodesic between the two subspaces on the Grassmann manifold to perform adaptation.

\noindent \textbf{SA} Subspace alignment \cite{DBLP:conf/iccv/FernandoHST13} - Using a single subspace representation for both domains, this approach aligns the subspaces via a linear transform.

\noindent \textbf{SDA} Subspace distribution alignment \cite{sun2015subspace} - In addition to aligning the source and target subspaces, this approach also aligns the actual distributions.

\noindent \textbf{PT} Domain adaptation via parallel transport on Grassmannian 
\cite{shrivastava2014unsupervised}- This approach considers the use of multiple subspaces on both the source and target, and performs parallel transport on the Grassmann manifold. Note, they report their results on only 8 of the 12 cases. 


A source (S) to target (T) domain adaptation is represented as S-T. For both SURF and Decaf representations, we repeated the adaptation experiments for all the $12$ cases. As described in Section \ref{sec:Methods}, we first represent the dataset via multiple subspaces, which resulted in 3 to 4 subspaces on an average for our datasets. We then align them using the proposed greedy approach. The results using the SURF and the Decaf features are reported in Table \ref{table:tab_Surf} and Table \ref{table:tab_decaf1} respectively. Some of the baseline results for Decaf features are from \cite{Luo2017DiscriminativeAG}. As it can be observed from the results, the proposed approach significantly outperforms the baseline cases. The subspace dimension $k$ was fixed at 45 across all the 12 settings, and the thresholds $\tau^s$ and $\tau^t$ were chosen from $[0.1,0.6]$. It is important to note that the three hyperparameters namely, the subspace dimension $k$, the error thresholds for source and target - $\tau^s$ and $\tau^t$ were tuned by grid search directly via the test error.


\begin{table}[t]
\begin{center}
\begingroup
\setlength{\tabcolsep}{10pt} 
\renewcommand{\arraystretch}{1.2}
\resizebox{0.75\columnwidth}{!}{%
\begin{tabular}{|c|c|c|c|c|}
\hline
\cellcolor{gray!25}\textbf{Domains} & \cellcolor{gray!25}\textbf{NA} & \cellcolor{gray!25}\textbf{SA} & \cellcolor{gray!25}\textbf{GFK} & \cellcolor{gray!25}\textbf{Proposed} \\
\hline
A-C & 78.54 & 79.61 & 80.32 & \textbf{81.38} \\

C-A & 87.05 & 87.06 & \textbf{87.27} & 86.95 \\

D-A & 75.89 & 73.49 & 75.27 &\textbf{ 79.54 }\\

A-D & 80.25 & 81.53 & 80.89 & \textbf{82.16} \\

W-A & 73.07 & 75.16 & 75.16 & \textbf{77.87} \\

A-W & 77.31 & \textbf{78.31} & 76.95 & 72.8 \\

W-C & 68.21 & 68.83 & 67.76 & \textbf{72.21}\\

C-W & 72.2 & 75.59 & 75.59 & \textbf{80.67 }\\

D-W & 97.97 & 98.98 & 98.98 & \textbf{99.66} \\

W-D & 98.98 & 100 & 100 & 100 \\

C-D & 80.89 & 80.25 & 83.44 & \textbf{87.26 }\\

D-C & 70.08 & 69.99 & 69.1 & \textbf{74.4} \\
\hline
\hline
Avg. & 80.12 & 80.73 & 80.90 & \textbf{82.98} \\

\hline
\end{tabular}%
}
\vspace{-15pt}
\endgroup
\end{center}
\caption{Classification accuracy on Office-Caltech dataset with decaf features}
\label{table:tab_decaf1}
\end{table}
%


\vspace{-2pt}
\section{Conclusion and Future Work}
\label{sec:conclusion}
We presented a novel domain adaptation method based on multiple subspace alignment. We proposed to represent data via a collection of low dimensional subspaces, and to subsequently align subspaces across source and target domains. Empirical studies on benchmark datasets were used to demonstrate the superiority of our approach in unsupervised DA for object recognition, when compared to existing subspace methods \cite{Gong2012GeodesicFK,DBLP:conf/iccv/FernandoHST13,sun2015subspace, shrivastava2014unsupervised}. Possible extensions to this work include exploring the optimal transport~\cite{villani2008optimal} techniques, in particular, the recent approaches designed for large-scale data~\cite{seguy2018large}, as an alternative for performing alignment between source and target representations, in order to alleviate the scalability issues with our approach. 


\small{
\bibliographystyle{IEEEbib}
\bibliography{main.bbl}}

\end{document}